# Employing Multimodal Machine Learning for Stress Detection

Rahee Walambe [1,2,*], Pranav Nayak [1], Ashmit Bhardwaj [1] and Ketan Kotecha [1,2,*]

[1] Symbiosis Institute of Technology, Symbiosis International (Deemed University), Pune-411215, India

[2] Symbiosis Centre for Applied Artificial Intelligence, Symbiosis International (Deemed University), Pune-411215, India

*Rahee Walambe (rahee.walambe@sitpune.edu.in) and Ketan Kotecha(director@sitpune.edu.in)

## Abstract

In the current information age, the human lifestyle has become more knowledge-oriented, leading to sedentary employment. This has given rise to a number of health and mental disorders. Mental wellness is one of the most neglected, however crucial, aspects of today's fast-paced world. Mental health issues can, both directly and indirectly, affect other sections of human physiology and impede an individual's day-to-day activities and performance. However, identifying the stress and finding the stress trend for an individual that may lead to serious mental ailments is challenging and involves multiple factors. Such identification can be achieved accurately by fusing these multiple modalities (due to various factors) arising from a person's behavioral patterns. Specific techniques are identified in the literature for this purpose; however, very few machine learning-based methods are proposed for such multimodal fusion tasks. In this work, a multimodal AI-based framework is proposed to monitor a person's working behavior and stress levels. We propose a methodology for efficiently detecting stress due to workload by concatenating heterogeneous raw sensor data streams (e.g., face expressions, posture, heart rate, computer interaction). This data can be securely stored and analyzed to understand and discover personalized unique behavioral patterns leading to mental strain and fatigue. The contribution of this work is twofold – namely, proposing a multimodal AI-based strategy for fusion to detect stress and its level and, secondly, identify a stress pattern over a period of time. We were able to achieve 96.09% accuracy on the test set in stress detection and classification. Further, we were able to reduce the stress scale prediction model loss to 0.036 using these modalities. This work can prove important for the community at large, specifically those working sedentary jobs, to monitor and identify stress levels, especially in current times of COVID-19.

## Introduction

The increase in the percentage of socio-economic category of knowledge workers has been on the phenomenal rise worldwide in the last few decades. In 2019, this socio-economic category surpassed 1 billion people, accounting for more than 30% of the world's total employed population [1]. With the majority of the world's working hours spent at a desk, employees' mental health becomes the most crucial issue. Workplaces are becoming more demanding than ever, requiring employees to deliver excellent results in the shortest amount of time. While this may appear acceptable at first, the health of the individuals suffers throughout prolonged working life, resulting in stress, worry, reduced productivity, and, in the worst-case scenario, depression. Workplaces have attempted to make employees' jobs more manageable but only so much can be done. To that end, it is critical to maintain the record and monitor people's mental health and performance over a period of time and take appropriate action as needed. Stress detection is a multimodal fusion problem involving various modalities of the data and can be solved using multimodal AI methods. In the following sub-sections the detailed literature review encompassing the multimodal fusion techniques in general applications, specific application of multimodal techniques in healthcare and stress detection, use of ML for stress detection are discussed. Various datasets that are useful for such tasks are also presented.

*Multimodal data and relevant applications*

The concept of combining data streams from several sources to achieve an outcome seems intuitive, yet there are several obstacles to overcome. Combining data from several sources, such as sensors, has proven to be more effective in forecasting outcomes. In [2], Ngiam et al. have examined multimodal deep learning. A novel approach to applying deep learning to different modalities like audio and video is proposed, and cross-modality feature learning is reported. This paper presents a method for learning enhanced representations for a single modality (e.g., video) from other modalities (e.g., audio and video) that are present throughout feature learning time. With the rise in low-powered sensors in wearable devices, the amount of data generated is enormous, but they are varied in discrete and continuous sampling rates. As a result, integrating them is challenging. In [3], Radu et al. propose a method for concatenating diverse sensor types utilizing four deep learning algorithms such as DNN and CNN. D. Lahat et al. [4] provide perspectives, guidelines, and ideas on multimodal data fusion approaches and their applications and techniques in multiple domains like health, biomedical and multisensory systems. In [5], Gros et al. present an in-depth analysis of the logic underpinning data fusion and discusses data fusion and multi-sensor integration approaches. Narkhede et al. [7] propose a method to detect gaseous emissions using multimodal data collected from gas sensors and thermal cameras. The fused model achieved 96% accuracy on the testing set instead of 82% on LSTM applied to sensor data and 93% on CNN applied to camera images for individual modalities. Cai et al. [8] review an innovative approach of using explainable AI on multimodal deep neural networks. This not only improves predictions owing to the usage of many modalities, but it also deviates from a neural network's black box decision-making and gives us insight into how the model arrived at any given result. Explainability also improves the model's trustworthiness and acceptability. There are multiple application domains where multimodal AI is employed, one of the most relevant being healthcare. Healthcare data is typically multimodal and has to be fused to obtain more meaningful outcomes.

*Multimodal data in healthcare*

The work related to multimodal data in healthcare is highly relevant. Before drawing any conclusions, medical specialists examine various images, data, and patient histories. So, if a machine learning algorithm is employed for decision making, having a mechanism for fusing multimodalities arising from various individual modalities becomes critical since any model is only as effective as the data it is trained on. Recently in 2019, Q. Cai et al. [8] tried to explore all the existing technologies and state-of-the-art methods used in the multimodal data healthcare industry. The USA, China, and Canada are the top three countries at the forefront of smart healthcare. In [9], Iakovidis et al. propose a semantic model to mine multimodal data is defined to be stored as feature spaces, easy to work with, train and test. On similar lines, [10] F. Wang achieved the same task implemented on top of the Hadoop framework, enabling parallelization. Collecting these datasets, let alone any model implementation on them, is a very long and tedious process. The few that are already present make them even more critical in the healthcare domain. Brain Tumor Image Segmentation Benchmark [11] is a multimodal dataset containing 3D brain MRI images used to detect brain tumors. It also contains a number of different approaches used to predict brain tumor presence and locations with accuracy scores and other metrics [11]. M. Cetin [12] researched Schizophrenia Classification used multimodal deep learning methodologies to predict the brain disease of a patient using fMRI and magnetoencephalography (MEG). They were able to achieve 85% accuracy using these modalities and ensemble neural networks of these two individual features. Radiology Objects in COntext (ROCO) [13] is a multimodal dataset to recognize the interaction of visual features and semantic links in radiological pictures. The goal is achieved by obtaining all image-caption pairings from PubMed Central, an open-access biomedical literature database, because captions represent visual content in its semantic context. Computer Tomography, Ultrasound, X-Ray, Fluoroscopy, Positron Emission Tomography, Mammography, Magnetic Resonance Imaging, and Angiography are among the medical imaging modalities included in the ROCO collection. Alzheimer's disease Neuroimaging Initiative (ADNI) [14] was collected to describe cross-sectional and longitudinal clinical assessments in healthy people, people with MCI, and people with mild Alzheimer's disease such that neuroimaging and chemical biomarker measurements may be evaluated. One of the exciting and most challenging areas in healthcare is stress detection. There are multiple approaches for stress detection using Machine learning and Artificial Intelligence.

*Stress detection using machine learning*

For stress detection, typically, questionnaires are created with the help of domain experts such as clinicians and psychologists. Such questionnaires are often used in research in the field of psychology to get insight into general working experiences and behavioral analysis of the participants. In areas where computing or the use of AI algorithms can be applied, the most commonly used modality is Electrocardiogram (ECG). However, using a single sensor modality to detect stress has certain limitations, such as less accuracy, more false positives/negatives, etc. Research from various fields shows the usage of different modalities and the potential use of sensors for estimating stress, mental and affective states, and the context in which they appear. Multiple modalities representing physical, neurophysiological, computer interactions, etc., can be fused to generate better outcomes. Since wearable sensors are getting more affordable and can be readily integrated into generic devices, such data can be generated and collected with ease. Saskia Koldijk et al. [15] have combined several modalities in a unique dataset with features like computer interactions, facial expressions, postures coordinates, and body sensors. In [16], Ahuja et al. have collected data of university students using a questionnaire and assigning certain weightage to these questions and then using various machine learning algorithms for predicting stress, with support vector machines yielding the highest accuracy (85%). In [17], Smets et al. have compared various machine learning algorithms for detecting mental stress on physiological responses in a controlled environment. They recruited 20 participants, conducted stress tests, and recorded data from two physiological sensors, wireless electrocardiography (ECG) sensor and NeXus 10 – MK II, to measure galvanic skin response GSR. In [18], J. Wijsman et al.. Measured physiological signals and features like skin conductance, Electrocardiogram, respiration, and surface electromyogram (sEMG) of the upper trapezius muscle wearable systems to predict stress in an office-like environment and reached an average accuracy rate of 74.5%. In[19], Picard et al. collected and analyzed physiological data of real-world driving tasks to determine stress levels. They found out that in most cases, driver's physiological data, i.e., heart rate metrics, skin conductance, etc., are closely correlated with driver stress level. In [20], Y. S. Can et al. have tried to perform continuous stress detection using unobtrusive wearable devices like Samsung S series devices and Empatica E4. They collected data from participants of an algorithmic programming contest for evaluation. An accuracy of 84% was achieved with Samsung S devices and multilayer perceptron, yielding the highest accuracy.

In [21], Mohd et al. aimed to present a novel approach for mental stress detection by using thermal imaging of the subject's face. They found a correlation between stress and blood flow in the face and have developed an automatic thermal face, Supraorbital, Periorbital, Maxillary, and Nostril Detection to estimate the person's internal state. In this work, we have considered the problem of identifying the stress of an individual based on various distinct different modalities using machine learning techniques. We considered the SWELL-KW dataset [15] for experimentation and demonstration of our multimodal fusion techniques.

*Multimodal datasets for stress detection*

Hence, in this work, we propose a strategy using multimodal artificial intelligence to classify mental stress and identify the scale of the stress. A dataset consisting of multimodal data called SWELL-KW [15] is used to validate and demonstrate our framework. This dataset is collected through the standard devices around any working individual to sense various modalities and utilize them for fusion using multimodal AI. SWELL-KW [15] is a powerful resource for accurately measuring sedentary jobs' work-associated mental stress. Most datasets regarding the stress monitoring domain have a single modality. However, the SWELL – KW dataset comprises four distinct modalities that, when combined, can be extremely useful for diagnosing stress and reliably predicting stress. The data was gathered as part of the SWELL Project by Wessel Kraaij et al. [22] and made publicly available in 2017. Since then, many forms of techniques have been applied to achieve state-of-the-art results in predicting stress based on the available modalities. The majority of stress detection research has focused on heart rate variability and related features as the data pairs well with related datasets like WESAD [23] and DREAMER [24]. S. Sriram Prakash et al. 2017 [25] implemented an SVM-classifier with RBF kernel to achieve 92.75% accuracy by employing only physiological signals available in the dataset. They also examined the individual features and their importance in predicting stress and concluded that the first stress indicator is Galvanic Skin Response and heart rate. In[26], Nkurikiyeyezu et al. have validated their model on SWELL-KW, which was trained using Advanced Trail Making Test [27], and achieved an accuracy of 99.25% using physiological data. In [28], S. Koldijk et al. focused on ranking the modalities to be more correlated to the prediction of stress and mental effort. The conclusion was that posture and facial expressions yielded the most valuable information. [29] S. Koldijk et al. showed us the visualizations of different modalities of the SWELL-KW dataset for better insights. SWELL-KW is a powerful resource for accurately measuring the work-related mental stress of sedentary jobs.

This work employs an artificial neural network (ANN) for feature extraction and early and late fusion-based techniques for multimodal data fusion, considering all four modalities. This approach is unique and has not been explored. In summary, the contribution of this work is threefold:

1. Implementating early and late fusion using machine learning to predict whether a person is stressed or not, given four specific modalities: computer interactions, body posture, facial features, and heart rate variability.

2. Applying transfer learning on early fusion features from the stress model to predict the NasaTLX score, which predicts the stress level on a scale of 0 to 100.

3. Providing a method to save the data from monitoring a person's mental state as the task load increases across the specific timeline.

The paper is organized as: Section II discusses the dataset employed and the methods applied for the multimodal fusion. Section III discusses the pipeline and workflow of the model. Section IV includes the findings and analysis of all of the predictions and assessment measures for each. Section V provides limitations of this work and the scope for future improvements. Section VI presents the conclusion.

## Materials and Methods

*DATASET*

To demonstrate our approach of multimodal fusion using machine learning, the SWELL Knowledge work dataset (SWELL-KW) [15] is used. This dataset was first presented in 2014 at 16th ACM International Conference on Multimodal Interaction by Koldijk et al. [15]. It is available in a publicly accessible repository [45]. The dataset was collected as a part of the research project wherein 25 subjects performed either traditional intelligence work or sedentary occupations. Making presentations, writing reports, reading emails, and researching information were all part of the experience. The participants' working environments were often exploited by the researchers, who exposed the subjects with stress-inducing stimuli such as email interruptions and time constraints. A total of 25 participants' data was produced. There were eight females and seventeen males in this study, with an average age of 25, comprising of Delft University of Technology students and TNO (the Netherlands Organization for applied scientific research) interns. Since they were workforce-ready, they had experience with large volumes of data and operating computers. Computer logs captured facial expression from camera recordings, body postures data points from a Kinect 3D sensor [36], heart rate variability, and skin conductance from sensors connected to the participant's body were all included in the dataset. The dataset contains raw, preprocessed and features extracted data, all readily available to work with. Validated questionnaires were administered to participants to assess their subjective interactions with task load, needed mental commitment, mood during these activities, and perceived stress. Participants were advised not to smoke or drink any caffeinated beverages three or four hours before the experiment because these are potential confounders. The experiment was classified into three blocks for the various stress conditions, with each session lasting approximately one hour. The dataset contains 3000+ examples that were used to train individual models. The ground truth labeling of whether the person was in a stressed state or not was provided along with the modalities' numerical values. Modalities collected are shown in Table 1.

**Table 1**: *Sample data details*

| Type | Available raw and preprocessed data | Available Features |
|------|-------------------------------------|--------------------|
| **Computer interactions** | uLog output & Parsed selection of data | Mouse (3) Keyboard (7) Applications (2) |
| **Facial expressions** | FaceReader output & Parsed data | Head orientation (3) Facial movements (10) Action Units (19) Emotion (8) |
| **Body postures** | Joint coordinates extracted with Kinect SDK & Angles of the upper body | Distance (1) Joint angles (10) Bone orientations (3x11) (as well as the study of the above for the amount of movement (44)) |
| **Physiology** | Data from Mobi | Heart rate (variability) (2) Skin conductance (1) |

a) *Computer logging*

The researchers used a background application on the users' computers. The application was a key-logging uLog [37] (version 3.2.5, by Noldus IT) for logging users' computer interactions. Table 2 shows some examples of computer logging data.

**Table 2**: *Sample examples for computer interaction data.*

b. *Facial features*

| Mouse Activity | Left Clicked | Right Clicked | Double Clicked | Wheel | … | Char Ratio | Error Ratio | Key |
|---|---|---|---|---|---|---|---|---|
| **0.0125** | 12 | 7 | 0 | 1 | … | 0.603774 | 0.216216 | |
| **0.401786** | 10 | 5 | 0 | 0 | … | 0.5 | 0.181818 | |
| **0.034188** | 0 | 0 | 0 | 0 | … | 0.7 | 0 | |
| **0.233333** | 34 | 12 | 0 | 0 | … | 0.605263 | 0.068966 | |
| **0.179916** | 37 | 17 | 0 | 2 | … | 0.875 | 0 | |

A high-resolution USB camera was deployed to record the participants' faces and upper bodies. The specifications of the camera were - iDS uEye UI-1490RE, 1152x768. To preserve the privacy of participants, no videos were made public. Researchers used a proprietary software called Noldus FaceReader [38] to interpret the data presented. Using deep learning and computer vision, this program analyses facial expressions in real-time. It provides details in the form of txt-logs containing information about facial expressions and emotions. FaceReader [38] app has over 30 functions, including head orientations, facial expressions, action units, and emotions. Table 3 shows some examples of facial expressions data.

*Table 3: Sample examples for facial expressions data.*

| Squality | Sneutral | Shappy | Ssad | Sangry | … | Ssurprised | Sscared |
|---|---|---|---|---|---|---|---|
| **0.944941** | 0.968862 | 0.023946 | 0.0013 | 0.016315 | … | 0.002024 | 0.001087 |
| **0.930303** | 0.88457 | 0.076952 | 0.001144 | 0.017392 | … | 0.002032 | 0.000651 |
| **0.933104** | 0.931965 | 0.031468 | 0.000371 | 0.023774 | … | 0.001722 | 0.001756 |
| **0.904466** | 0.806947 | 0.105516 | 0.006459 | 0.009809 | … | 0.001563 | 0.000441 |
| **0.929025** | 0.951412 | 0.028358 | 0.001095 | 0.01813 | … | 0.001309 | 0.003466 |

c. *Body Posture*

A per-time frame analysis of the participant's body orientation was included in the data sets. They acquire the coordinates of all the joints by fitting the Kinnect [36] skeletal model in this manner. These CSV files contain all of the coordinates necessary to determine angles between upper-body joints and bones. The dataset also includes upper body bone orientations with timestamps relative to the x, y, and z axes. Over 90 characteristics were included in the posture data. Table 4 provides few samples of observations on body posture.

Table 4: Sample examples for body posture data.

| Avg Depth | Left Shoulder Angle avg | … | Right Shoulder Angle avg | Lean Angle avg |
|---|---|---|---|---|
| | | | | |
| **2102.597393** | -116.055931 | … | 115.017758 | 92.340895 |
| **2099.725525** | -116.301605 | … | 115.986636 | 92.083385 |
| **2102.365778** | -115.963089 | … | 114.073054 | 92.38169 |
| **2104.116968** | -115.62963 | … | 113.972465 | 92.428905 |
| **2105.284007** | -116.359699 | … | 111.538437 | 92.461537 |

*d. Body sensors*

The ECG was recorded using a Mobi unit (TMSI [39]) with self-adhesive electrodes. The recording software Portilab2 [40] was created with some preprocessing. Skin conductance was measured using Mobi and finger electrodes.

Out of these 3000 + samples, due to failure in capturing a reading at any particular moment for *all* modalities, only 956 examples that reported all three modalities correctly, were eventually used to train the final model on 70-30 train-test split in this work. The archive contains over 900 documents, containing both raw and *s*tructured records. Since some of the functions were not labeled and had many missed values, we had to merge and preprocess several files with each modality. We closely examined these files before selecting and sorting different files and merging them to create a single data file. Python was used for all of the data preprocessing activities.

## METHODS

**Artificial neural network**

A neural network [30] is a layer-by-layer connection of neurons that attempts to replicate the functioning of the human brain. The first layer of a neural network is the input layer, and the last is the predicted output layer. The hidden layers between the input and output layers take the output of the last layer's neurons as input and return some output after a mathematical calculation. Each layer is added in a sequential order, with the previous layer's output serving as the input for the next layer.

*Dropout*

Some selected neurons are disregarded during the training process and are not included in the computation of the output or in the backpropagation [31]. Since each neuron is trained on a specific collection of examples, this helps us avoid overfitting. The neurons are dropped out to differ in each epoch and are chosen at random. Here, a dropout [32] rate of 0.5 is used, which means that 50% of neurons would be ignored at each step.

*Activation functions*

The representation power of a deep N.N. is due to its non-linear activation functions.

Sigmoid: Sigmoid activation is implemented at the output layer. The focus is on biclass classification (stressed or not). Hence sigmoid activation is the best fit since it predicts the probability as an output.

The mathematical formula for this is shown in Equation 1.

$$f(x) = \frac{1}{1+e^{-x}} \qquad (1)$$

*ReLU*: Rectified Linear Unit is an activation function that increases linearly for positive inputs and outputs zero for negative inputs. The formula for the ReLU function is seen in equation 2. ReLU is used for the model's hidden layers.

$$f(x) = max(0, x) \qquad (2)$$

*Loss functions*

*Binary Cross-Entropy*: The final layer generates output that is compared to the ground reality, and a loss function is used to quantify the error, which is then back propagated [29] to train individual neurons' formulae for improved results. The formula for binary-cross entropy [33] is as shown in equation 3, where $\hat{y}_i$ is the i-th value predicted by the model, $y_i$ is the corresponding actual value, and the output size is the number of scalar values in the model output.

$$loss = -\frac{1}{output\ size}\sum_{i=1}^{output\ size} Y_i * \log \hat{y}_i + (1 - y_i) * \log(1 - \hat{y}_i) \qquad (3)$$

*Root Mean Squared Error*: On regression model predictions, Root Mean Square Error [34] is the most suitable evaluation metric. The RMSE is computed as shown in equation 4, where N is the number of examples, $\hat{x}_i$ is the value predicted by the model, and $x_i$ is the actual value or observation.

$$loss = \sqrt{\frac{\sum_{i=1}^{N}(x_i - \hat{x}_i)^2}{N}} \qquad (4)$$

**Early fusion**

Early fusion techniques incorporate various modalities by constructing a joint representation of input features. The final prediction can be expressed as seen in equation 5, where concatenation indicates concurrently represented modality features. Since only one model is used, the training procedure is simple. It often requires highly engineered and preprocessed features from several modalities in order for them to align well or have similar meanings [35].

$$p = h(v_1, v_2, \ldots \ldots v_m) \qquad (5)$$

$h$ is used to denote a single model, $v$ stands for input features from multiple modalities, $P$ is final prediction. In layman's terms, early fusion occurs as the modalities are merged or feature are mapped *before* attempting to classify them.

**Late Fusion**

Late fusion employs a fusion technique to combine decision values from individual modalities [35]. Assume model hi is used on modality i (i = 1, .., M,) the final prediction is shown in equation 6.

$$p = F(h_1(v_1), h_2(v_2) \ldots \ldots h_m(v_m))  \qquad (6)$$

The late fusion method admits the employment of several models on various modalities, providing for greater flexibility. Because the predictions are created independently, it is easier to deal with a missing modality. In layman's terms, late fusion entails classifying outcomes through individual modalities before integrating the model predictions to characterize the final production.

**NASA-TLX**

The Nasa Task Load Index [41] is a measurement of the workload of any particular job. It was developed over three years by NASA's Ames Research Center's Human Performance Group, which used more than 40 laboratory simulations. It considers all aspects of a job, including mental need, physical demand, temporal demand, efficiency, effort, and frustration. NasaTLX scores range from 0 to 100, with 0 representing rest mode or no work requirement or effort at all and 100 representing a task that requires complete efforts, both mental and physical.

**MODEL**

Our system design maps three individual neural network architecture components to predict status based on body orientation, facial expression, and keystroke dynamics. Vanilla Neural Network is employed as an individual network for all three modalities as the data was present in numeric form. The final layer of each modality neural network can then be linked with other neural networks to form an ensemble neural network architecture [42]. Our model flow uses a Hard Parameter Multi-task Learning, wherein the model has common layers that split into task-specific layers further. This simply indicated that the feature maps are used to transform a large number of individual modalities' features into a small number of each and used that as an input to our ensemble neural network. The model hyperparameters were selected after experimenting with different combinations, and the model which outperformed in training and testing phases has been explained below. The model was trained using cross-validation [46] to prevent overfitting. The system was trained on approximately 3,000 examples over 200 epochs. Each epoch took, on average, one minute to train and the model as a whole took 16 hours - 10 hours to train individual models and 3 hours each to train the stress classifier and NASA-TLX regressor.
    A) STRESS CLASSIFIER

Each of the individual neural network layers has been equipped with a dropout layer. This helps us prevent overfitting the data as several columns might contain irrelevant information or might not be as useful as others. ReLU activation function is used in hidden layers as it demonstrated the best results in all our findings. Sigmoid activation is used for the output layer, giving us the probability of whether a person is stressed or not (bi class prediction). In the case of the keystroke dynamics neural network, the model's number of parameters was not too high, so a simple neural network with one hidden layer was enough to produce good results. However, in the case of models such as facial expressions and body posture, there was a need for a more complex neural network due to the higher number of features in each modality. For skin conductance and heart rate variability metric, two features and one feature are present respectively. Hence different neural network architecture was not necessary. Instead when the output for the individual models was generated, these three features(namely, skin conductance(2), heart rate variability(1)) were provided as the input to our next and final neural network, which was also equipped with a dropout layer and used ReLU and sigmoid layer on the hidden and output layer, respectively. All the neural network models were trained on the binary-cross entropy loss function. The results for both early and late fusion were compared.

In early fusion, output of the last hidden layer is used as an input to the combination neural network. Figure 1 depicts the construction of individual architectures and the use of the early fusion technique to predict stress and NasaTLX score. In the late fusion prediction, the probabilities of individual models for stress are provided as an input to the final neural network. Figure 2 shows the architecture for the late fusion technique to classify whether a person is stressed or not.

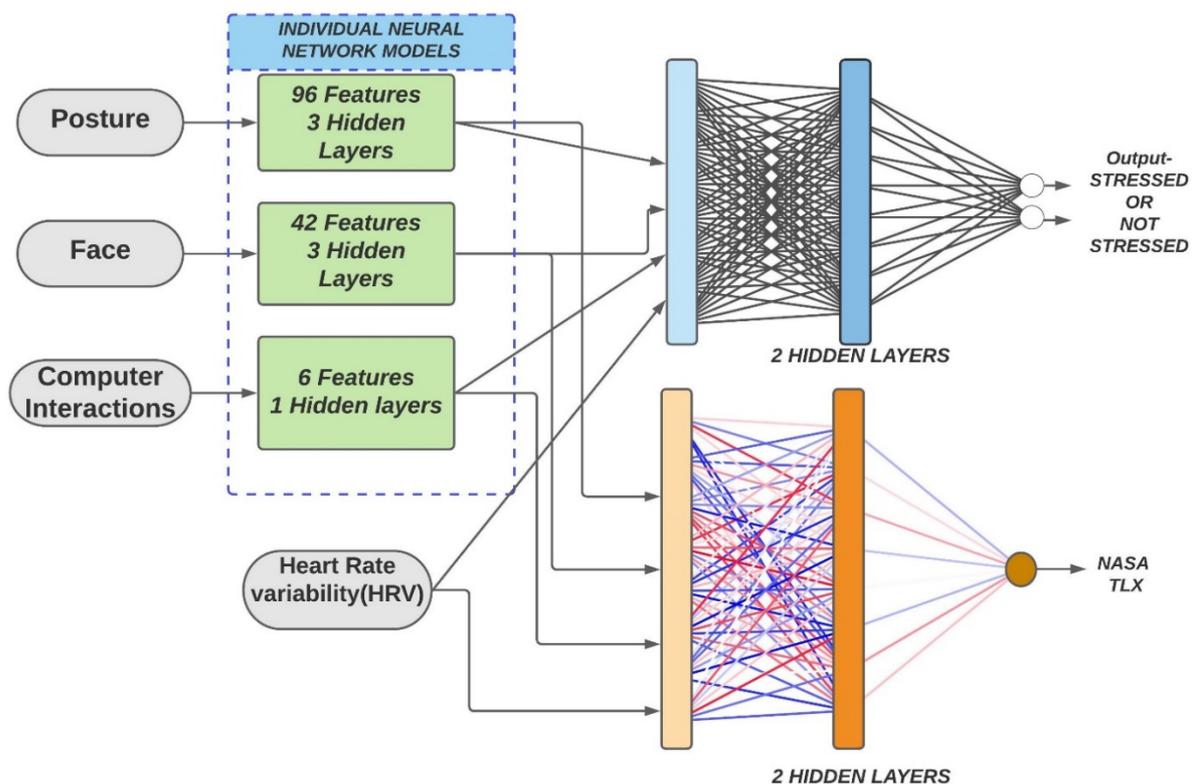

*Figure 1: Model architecture and workflow for Early Fusion and NasaTLX Prediction.*

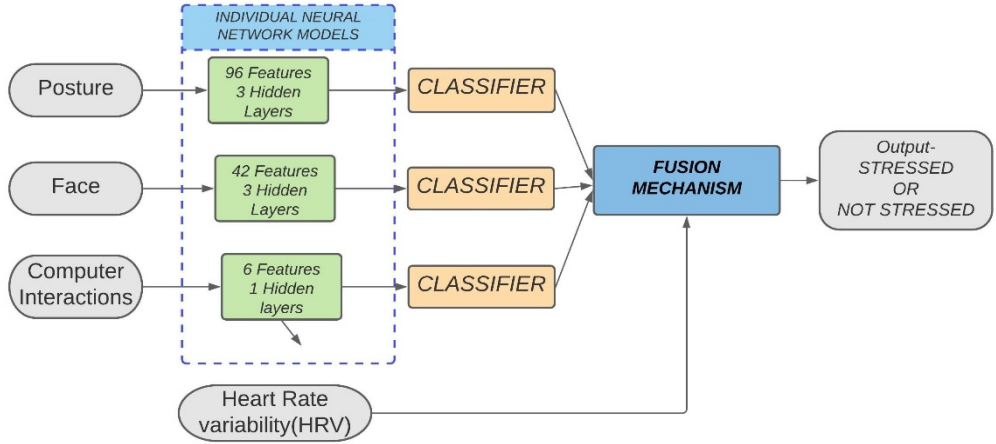

Figure 2: *Model architecture and workflow for the Late Fusion technique.*

### B) NASA-TLX REGRESSION MODEL

The same feature map used to predict stress with early fusion was a good predictor of Nasa-TLX scores. With this, it can be observed that the feature maps our original input as an indicator of stress. Similar to the stress detection, a 2-hidden layers network was designed from the output of individual neural networks to make an ensemble neural network regression model. RMS Error was used as the loss function for the same. This is a transfer learning solution carried out as the same model feature map was used for a prediction of different but related entity. NasaTLX predictions were also carried out with late fusion model but the results were not significant. This can be attributed to the fact that late fusion model had only a few features which were not able to scale the features to an extent that early fusion could.

## Results and Discussion

The prediction and analysis tasks were divided into two streams - prediction of NASA Task Load Index using regression model and predicting whether the user is stressed or not using neural networks classification.

**A. NASA-TLX**
We achieved an RMSE of 0.047 on the training set and 0.036 on the test set in neural network predictions. The better model performance on the test set can be attributed to the dropout layer, which is at its optimum capacity only during the test phase. Figure 3 shows how loss varies with increasing epochs during the training phase.
Note that the NasaTLX score scale is 0-100, so an average loss of 0.036 on this scale is minimal.

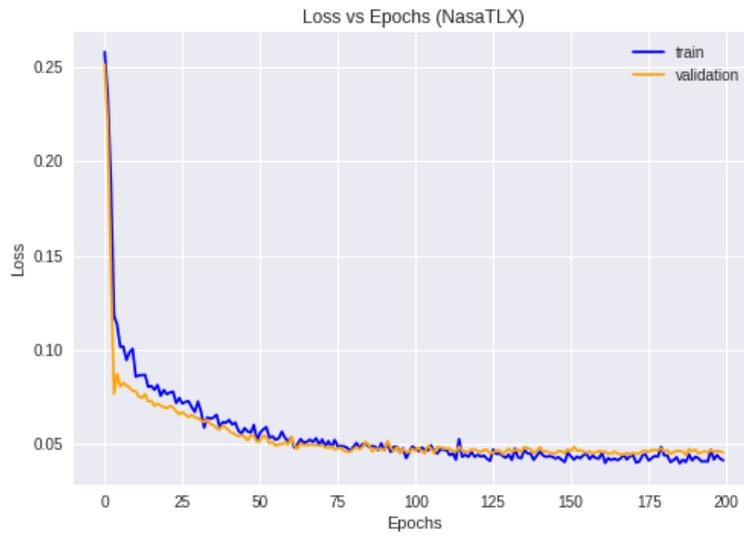

*Figure 3: Loss vs. Epochs – NasaTLX model.*

## B. STRESS DETECTION

The metrics of the individual stress classification model and the ensemble neural network architecture can be found in table 5. As seen, body posture is the best indicator of stress, giving an accuracy of 77%.

Table 5: Evaluation metrics for individual models

| Modality | Accuracy (%) | Precision | Recall | F1 Score |
|---|---|---|---|---|
| *Body Posture* | 77.56 | 84.45 | 76.01 | 78.03 |
| *Facial Expressions* | 74.05 | 82.05 | 74.02 | 71.02 |
| *Keystroke Dynamics* | 71.33 | 72.45 | 68.02 | 71.01 |

Both, early fusion and late fusion technique was used on these three models to form the main neural network for final predictions. These early and late fusion outputs were also added with three heart rate variability and skin conductance features. Figure 4 shows the training and validation accuracy plots for both late and early fusion models demonstrating the superiority in accuracy and early convergence of the early fusion model. Figure 5 shows the loss charts for both early and later fusion. It can be seen that the loss goes on decreasing with increase in epochs and for early fusion the loss is even smaller as compared to the late fusion model. Figure 6 and Figure 7 present the confusion matrices for both the respective models. The false positives and false negatives for the early fusion model are comparatively lesser than the late fusion model proving the better performance of the early fusion model. Figure 8 demonstrates the residual plot for the predictions made by NASA-TLX regression model on the test set. As clearly visible, most of the predictions lie between ±0.5 with only a few outliers going out of ± 2 range. Note that the score is in the range of 0-100 so a decimal error is relatively affordable. Figure 9 shows the ROC curve demonstrating the classifier performance at every threshold.

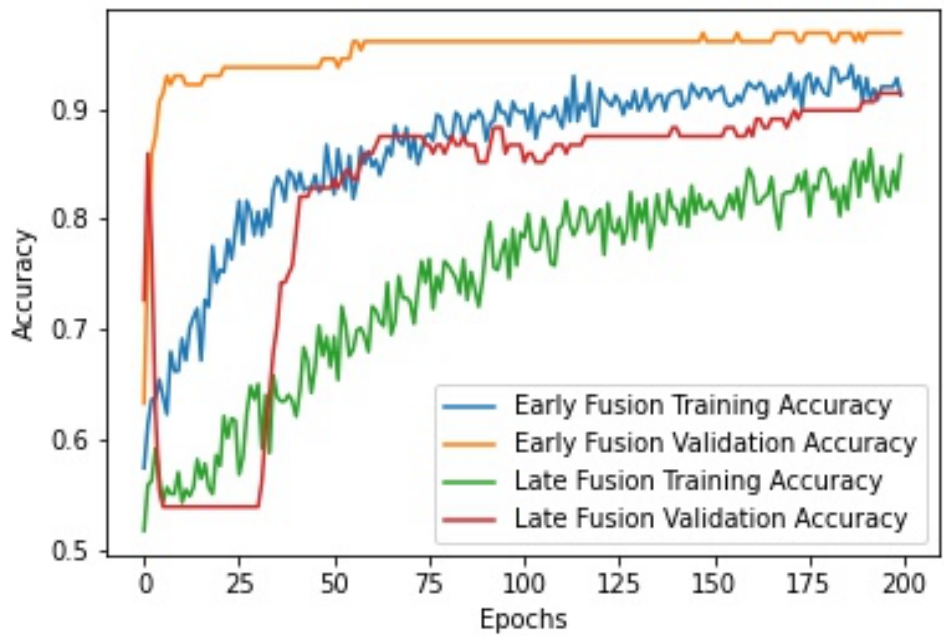

Figure 4: Accuracy chart for Early and Late Fusion

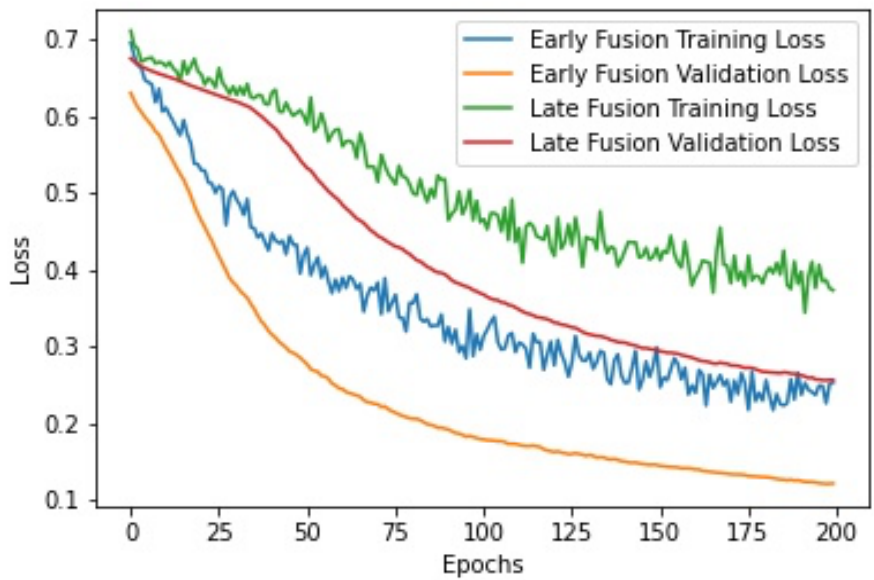

Figure 5: Loss plot for Early and Late Fusion

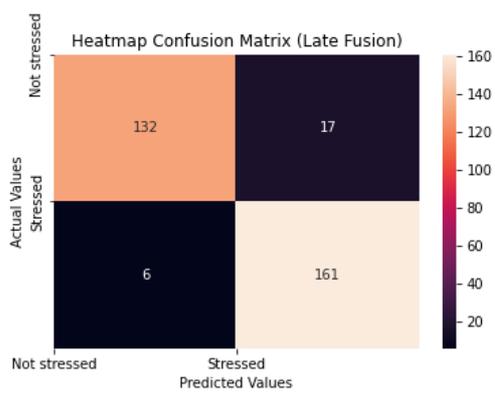 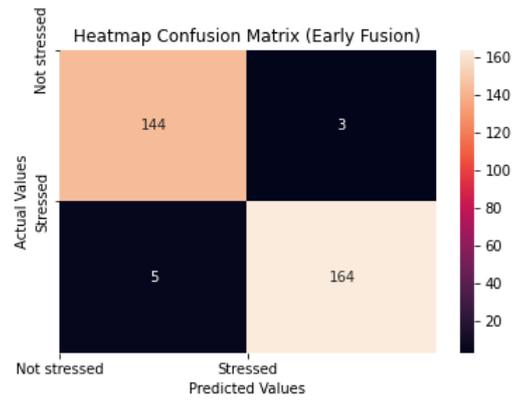

*Figure 6: Confusion matrix for Late Fusion.*   *Figure 7: Confusion matrix for Early Fusion.*

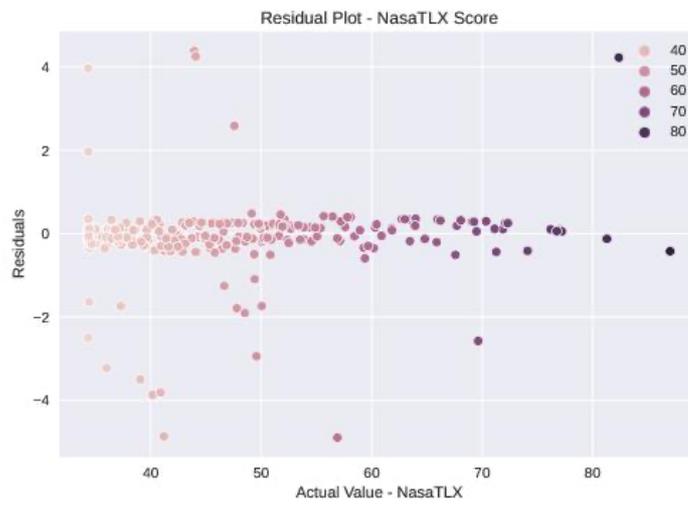

Figure 8: Residual plot for NasaTLX prediction model

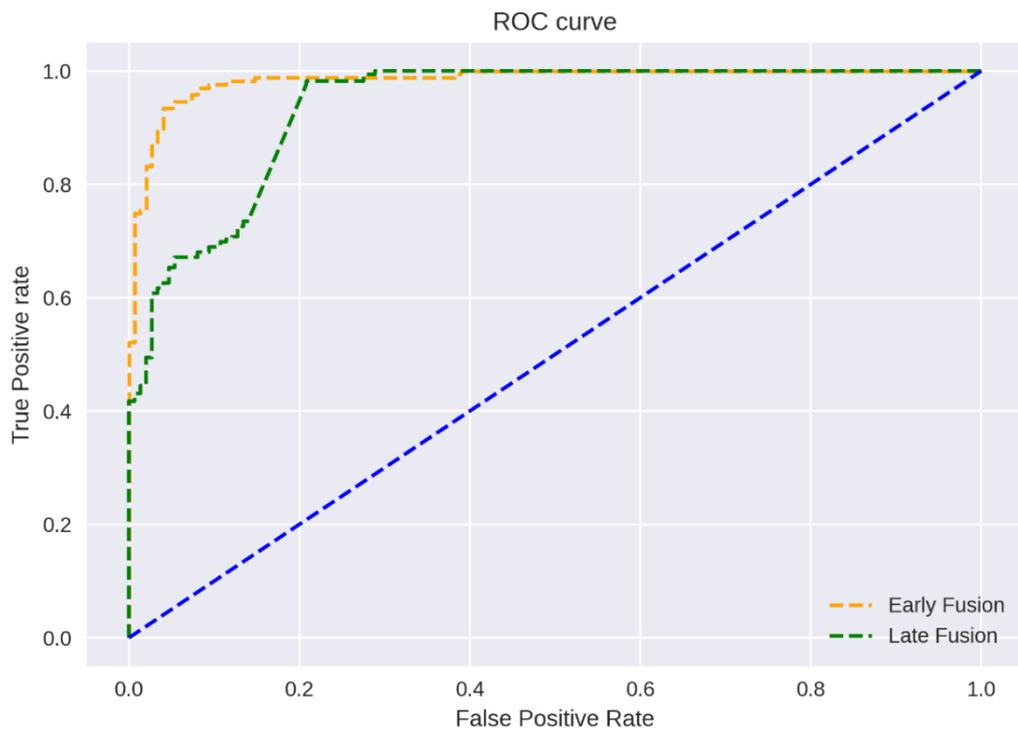

Figure 9: ROC Curve for the Early and Late Fusion models

Table 6 shows the evaluation metrics for each of the final models using early and late fusion respectively.

Table 6: A comparison of metrics for early vs. late fusion models.

| Fused Model | Accuracy (%) | Precision | Recall | F1 Score |
|---|---|---|---|---|
| *Late fusion* | 90.45 | 0.91 | 0.90 | 0.90 |
| *Early fusion* | 96.67 | 0.95 | 0.95 | 0.95 |

Table 7: Other research work on SWELL-KW dataset; the modalities used and accuracy scores.

| Research Model | Modalities used | Accuracy |
|---|---|---|
| **SVM-classifier with RBF kernel[25]** | Heart Rate Variability and Physiological Data | 92.75% |
| **Fast-GRNN[43]** | Heart Rate Variability and Physiological Data | 87.87% |
| **Support Vector Machine[28]** | Heart Rate Variability, Computer interactions, body posture and facial features | 90% |
| **Active Bayesian Learning[44]** | Heart Rate Variability and Physiological Data | 91.92% |
| **Our Model** | Heart Rate Variability, Computer interactions, body posture and facial features (with early fusion) | 96.67% |

**Comparison with other works using the SWELL-KW dataset:**
A number of approaches for detecting the stress are reported on SWELL-KW Dataset. These approaches employed subset of the available modalities and their accuracy scores. In [27], similar stress detection experiment on different database and using different metrics for stress measurement are carried out. Hence the results from [27] are not included. However, it can be seen that using the subset of modalities and implementing the various machine learning models such as SVM [25,28], Fast GRNN [22] and Active Bayesian Learning [43] underperform our model with early fusion. Table 7 summarizes the research models used for the same dataset by other researchers, their modalities used and accuracies achieved. Our model outperforms all of the present models, achieving a state-of-the-art accuracy score, which can be credited to the use of multimodal fusion techniques with an ensemble neural network model. With all these evaluations, it becomes evident that early fusion performs better than late fusion techniques. We can attribute this to the fact that many features of individual modalities are better mapped by early fusion, which ameliorates our final result. Finally, the real-time prediction is demonstrated in Figure 8. Figure 10 consists of the plot showing "orange" whenever the state of the user is stressed and "blue" whenever they are relaxed. The graph clearly shows that when workload increases, the person starts to get stressed when pursued too long. Some orange dots in the middle might be false indicators of stress, so the administrator monitoring this might not notify the user if the stress levels are not high and persistent for long periods. Figure 9 shows the ROC curve demonstrating the classifier performance at every threshold, which clearly indicates that early fusion has a higher area under curve, hence, better predictions which is consistent with other evaluation metrics.

**Limitations**

This research provides a high accuracy for stress classification using multimodal AI data fusion techniques, but there are a few limitations. The model works best with availability of more modalities. It may work with lesser modalities but the idea is to have samples from as many modalities representing stress as possible leading to better accuracies and more importantly lesser false positives or false negatives. The state of our model, as it stands, needs all the input parameters to produce the results so, future work may include an extension of this research on multimodal co-learning where the study can be carried out to understand the robustness of the multimodal model in the absence of one more modalities at test /train time. This will benefit the users who cannot provide all the modalities.

## Conclusions

This paper investigates new ways to leverage the SWELL-KW dataset to predict stress levels and task load based on the modalities provided in the dataset. We used different multimodal fusion algorithms for the predictions and evaluated them to compare and report the best one. The early fusion model showed the best results on stress classification. It also showed better results than multiple linear regression models for predictions of task load. Finally, we showed how the data could be stored according to the timeline, which clearly shows that a prolonged increase in task load leads to stress. The input modalities can be easily replicated using simple resources around any knowledge worker who makes this set easy to use in any environment. The factor of stress affecting any person should not be ignored for too long as it causes health issues, both mental and physical. Furthermore, using the power of artificial intelligence in healthcare that goes beyond our supervision will go a long way.

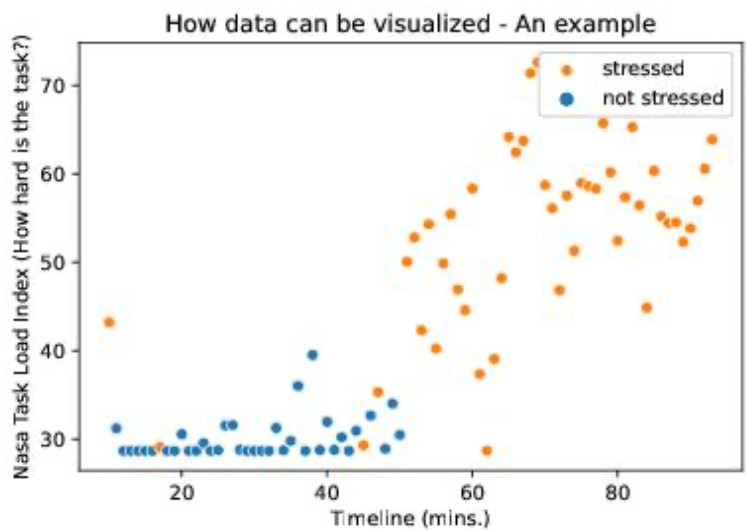

Figure 10: Timeline showing a person's state as the task load increases.

## Data Availability

The SWELL KW dataset can be accessed at [15].

## Conflicts of Interest

The author(s) declare(s) that there is no conflict of interest regarding the publication of this paper.

## Funding Statement

This work is funded by Symbiosis International University, Pune, India under the research support fund.